\documentclass[conference]{IEEEtran}
\ifCLASSINFOpdf
   \usepackage[pdftex]{graphicx}
\else
\fi

\usepackage{amsmath}
\DeclareMathOperator{\sign}{sign}

\begin{document}
%
\title{Cluster Metric Sensitivity to Irrelevant Features}
%
%
%


\author{\IEEEauthorblockN{Miles McCrory}
\IEEEauthorblockA{\textit{Data Science Department} \\
\textit{National Physical Laboratory}\\
Hampton Road, Teddington, United Kingdom \\
m.mccrory@npl.co.uk}
\and
\IEEEauthorblockN{Spencer A. Thomas}
\IEEEauthorblockA{\textit{Data Science Department} \\
\textit{National Physical Laboratory}\\
Hampton Road, Teddington, United Kingdom \\
spencer.thomas@npl.co.uk}
}

\maketitle

\begin{abstract}
Clustering algorithms are used extensively in data analysis for data exploration and discovery. Technological advancements lead to continually growth of data in terms of volume, dimensionality and complexity. This provides great opportunities in data analytics as the data can be interrogated for many different purposes. This however leads challenges, such as identification of relevant features for a given task. In supervised tasks, one can utilise a number of methods to optimise the input features for the task objective (e.g. classification accuracy). In unsupervised problems, such tools are not readily available, in part due to an inability to quantify feature relevance in unlabeled tasks. In this paper, we investigate the sensitivity of clustering performance noisy uncorrelated variables iteratively added to baseline datasets with well defined clusters. The clustering quality is evaluated using labeled and unlabelled metrics, covering a range of dimensionalities in the baseline data to understand the impact of irrelevant features on clustering popular metrics. We show how different types of irrelevant variables can impact the outcome of a clustering result from $k$-means in different ways. We observe a resilience to very high proportions of irrelevant features for adjusted rand index (ARI) and normalised mutual information (NMI) when the irrelevant features are Gaussian distributed. For Uniformly distributed irrelevant features, we notice the resilience of ARI and NMI is dependent on the dimensionality of the data and exhibits tipping points between high scores and near zero. Our results show that the Silhouette Coefficient and the Davies-Bouldin score are the most sensitive to irrelevant added features exhibiting large changes in score for comparably low proportions of irrelevant features regardless of underlying distribution or data scaling. As such the Silhouette Coefficient and the Davies-Bouldin score are good candidates for optimising feature selection in unsupervised clustering tasks. 
Finally, we observe that standardizing and mean centering the data prior to clustering removes the discrepancies between Gaussian and Uniformly distributed irrelevant features and in general reduces variability in metrics between repeated cluster runs. 
\end{abstract}


%
\IEEEpeerreviewmaketitle

\section{Introduction}
Clustering is an important unsupervised machine learning method and can be applied in pattern recognition, image segmentation and data mining problems, \cite{kogan_survey_2006}. Clustering algorithms group similar points together based on a measure of distance or similarity of the features within the points, such as the Euclidean distance between their values or attributes. 

In many applications of clustering, the ground truth labels are not available and the algorithms are used in a data exploration methodology or pattern recognition. Therefore, in order to have confidence in the resulting groupings, we must understand the impact of input data on the clustering results. Datasets are increasing in volume, dimensionality and complexity, often with distinct possible clustering tasks of interest based on subsets of the data. As such, for a given task of interest, the input data can contain redundant or irrelevant variables for that specific task. Feature selection methods can help identify and remove irrelevant features and have been widely used in supervised learning tasks, \cite{papaioannou_parallel_2023, wang_feature_2016, pudjihartono_review_2022,zabalza_novel_2016,hira_review_2015,saeys_review_2007}. However, these require ground truth labels in order to assess the impact of removing a feature for the learning task, such as increased accuracy (classification) or reduced error (regression), \cite{dash_feature_1997}. For unsupervised tasks, such as clustering, these labels are not available, nor is there a clear objective to assess feature selection. Therefore it is important to understand how sensitive clustering metrics are to the inclusion of irrelevant features for a task when assessing assigned clusters, \cite{houle_can_2010}. Understanding these sensitivities can potentially identify candidate metrics for performing unsupervised feature selection, by providing a suitable objective to optimise.

In practice, we are unlikely to know {\em a priori} if variables in the data are uncorrelated to a given task. Typically exploratory data analysis is used to identify patterns and key features in the data that are then used to inform feature selection and down stream analysis. However, many real word problems are highly complex and the relationship between a feature and the target of interest may be unknown. Furthermore, individual inputs alone may not correlate to the task or output, but there may exist a nonlinear combination of inputs that describes the task well. Similarly, random variables can negatively impact clustering as the random distances will mask any useful information in the data, \cite{leonard_kaufman_peter_j_rousseeuw_introduction_1990}. From a practical perspective, we may not know which of the variables are or are not important, but we would like our evaluation metrics to reflect if the data contains non informative features. This at least provides confidence in our interpretation of the metric values as `good' or `bad'. Additionally, this can enable feature selection through optimisation of the metric, thus providing an automated way to remove redundant or irrelevant features in an unsupervised data-driven way.

There exists a range of clustering algorithms that can be used effectively for specific tasks and applications, \cite{cheng2019local}. k-means is a centroid based algorithm and is one of the most popular methods as it is quick to run and easy to implement, \cite{kogan_survey_2006, arthur_k-means_2007}. 
The k-means algorithm minimise the distance of points within clusters, the within-cluster sum of squares, but maximise the distance of points between clusters, the between-cluster sum of squares. The algorithm iterates through a two step process, firstly assigning each point to the nearest clusters, and then updating the calculated centroids, \cite{MacKayInfromationTheory2003}. 
The k-means algorithm requires the user to state the number of clusters {\em a priori}. If this is not known, the number $k$ of clusters can be estimated using the so called `elbow plots', such as the inflection point of total sum squared distances to cluster centroids as a function of $k$, \cite{yuan_research_2019}, maximising metrics such as the Silhouette Coefficient as a function of $k$, or directly using data-driven embedding, \cite{Thomas2022cibcb}. Density based clustering methods such as DBSCAN, \cite{ester_density-based_1996} avoid the need to know the number of clusters {\em a priori}, however, they do require the selection of a minimum number of points and radius parameter. This replaces one unknown parameter with two, which additionally may not be easy or intuitive to optimise. Due to the cluster number being a comparatively simple and interpretable parameter to optimise, and the algorithms ubiquitous use in clustering problems, in this work we focus on k-means. 

In this work we investigate the sensitivity of the k-means clustering algorithm to increasing levels of random variables in the input data in order to study the impact on clustering performance. This is effectively increasing the level of impact of the so called 'curse of dimensionality', \cite{houle_can_2010}. We conduct experiments on datasets with ground truth labels to verify useful metrics for practical applications where such information is not available. By artificially adding increasing amounts of random variables to input data we can determine the impact of these irrelevant features relative to the informative features in the data. We define irrelevant here as features that are uncorrelated with the cluster label, modelled as randomly generated values, that are added across all cluster groups. Our results cover several datasets of different dimensions and we monitor the ratio of random variables to informative features, evaluating clustering performance using several metrics. We investigate Gaussian and uniformly generated random values, as well the effect of scaling the data. This work could easily be extended to other types of clustering algorithms and different distributions for sampling random numbers. To the best of our knowledge this has not been investigated previously with the literature surrounding clustering sensitivity examining the internal parameters of clustering algorithms, i.e ablation studies or (hyper) parameter tuning, \cite{peng2022clustering,krzak2019benchmark, peng2020average, roux_cluster-based_2021}, or
sensitivity analysis, \cite{kristiansen2017sensitivity, hajnal_sensitivity_1996, olukanmi_sensitivity_2017}.

\section{Problem Formulation}


\subsection{Data}
\label{sec:data}
The datasets used in the experiments are called the Dimsets datasets, \cite{DIMsets}. There are four different datasets available each with a different number of dimensions, $D =\left( 32, 64, 128, 256 \right)$, referred to as Dim-$D$ where $D$ is the dimensionality. The datasets all have $1024$ data points and $16$ clusters each made up of $64$ data points. These clusters are generated to have a Gaussian distribution with clusters that are well separated in all dimensions. All of these datasets have associated ground truths and the initial clustering metric scores for these dataset can be used as a baseline reference point for all other datasets when  irrelevant features are iteratively added.

\subsection{Evaluation Metrics}
\label{sec:metrics}
There are various metrics that can be used to measure the performance of clustering algorithms. Clustering metrics assess performance in two main ways; either by comparison of predicted labels to a ground truth or by measuring spatial distances within and between clusters. In this study we have the associated ground truth labels and evaluate the clustering results with this information. 

Normalised Mutual Information (NMI) compares a clustering outcome ($X$) to the ground truth clustering labels ($Y$) defined as
\begin{equation}
\label{eq:NMI}
\textrm{NMI} = \frac{\textrm{MI}\left(X,Y\right)}{\textrm{mean}\left(H\left(X\right), H\left(Y\right)\right)}~,
\end{equation}
where $\textrm{MI}()$ is the mutual information, $H(z)$ is the entropy of $z$. 
This has an upper bound of 1 indicating perfect clustering assignment and a lower limit of zero for incorrect clustering results, \cite{scikit-learn}. 

The Rand Index (RI), \cite{rand_objective_1971} is a measure of similarity between two sets of data groupings, in our case this is the similarity between the cluster results and the ground truth labels and defined as
\begin{equation}
\label{eq:RI}
\textrm{RI} = \frac{a+b}{N}
\end{equation}
where $a$ and $b$ are the number of true positives and true negatives respectively. $N$ is the total number of points in the data. As we are investigating the impact of random variables on clustering results, we use the adjusted Rand Index (ARI) 
\begin{equation}
\label{eq:ARI}
\textrm{ARI} = \frac{\textrm{RI}-E\left[\textrm{RI}\right]}{\max(RI)-E\left[\textrm{RI}\right]}~,
\end{equation}
where $E\left[~\right]$ is the expectation value due to the random elements in the k-means initialisation leading to differing $a$ and $b$ across runs. This formulation ensures that random labels will have scores near zero. This has an upper bound of 1 indicating perfect clustering assignment.  

The Silhouette Coefficient ($\textrm{S}$) measures the clustering results assuming the desired outcome is dense and separated clusters. It is defined using the mean distance between a point and all points within the same cluster (${d_w}$) and the mean distance between a point and all other points in the nearest cluster (${d_n}$)
\begin{equation}
\label{eq:sil}
\textrm{S} = \frac{{d_n} - {d_w}}{ \max \left( {d_n}, {d_w} \right)}~.
\end{equation}
This has a lower limit of -1, sparse and overlapping clusters, and 1 for dense well separated clustering.


The Davies-Bouldin Index ($\textrm{DB}$), \cite{davies_cluster_1979} compares the similarity of each cluster with the next most similar cluster within the dataset, averaged over all $k$ clusters. The $\textrm{DB}$ is calculated using the average distance of all points to the centroid ($\delta$) within each cluster, and the distance between centroids for pairs of clusters ($\Delta_{ij}$),
\begin{equation}
\label{eq:DB}
\textrm{DB} = \frac{1}{k} \sum_{i=1}^k  
\max_{{j \neq i}}
\left( \frac{\delta_i + \delta_j}{\Delta_{ij}} \right)
\end{equation}
The perfect Davies-Bouldin score is $0$ which means clusters are well separated, clearly defined and dense. There is no upper bound for the Davies-Bouldin metric but a higher scores means clusters are poorly defined and overlap.

\subsection{Experiments}
\begin{figure}[t]
    \centering
    \includegraphics[width=0.45\textwidth]{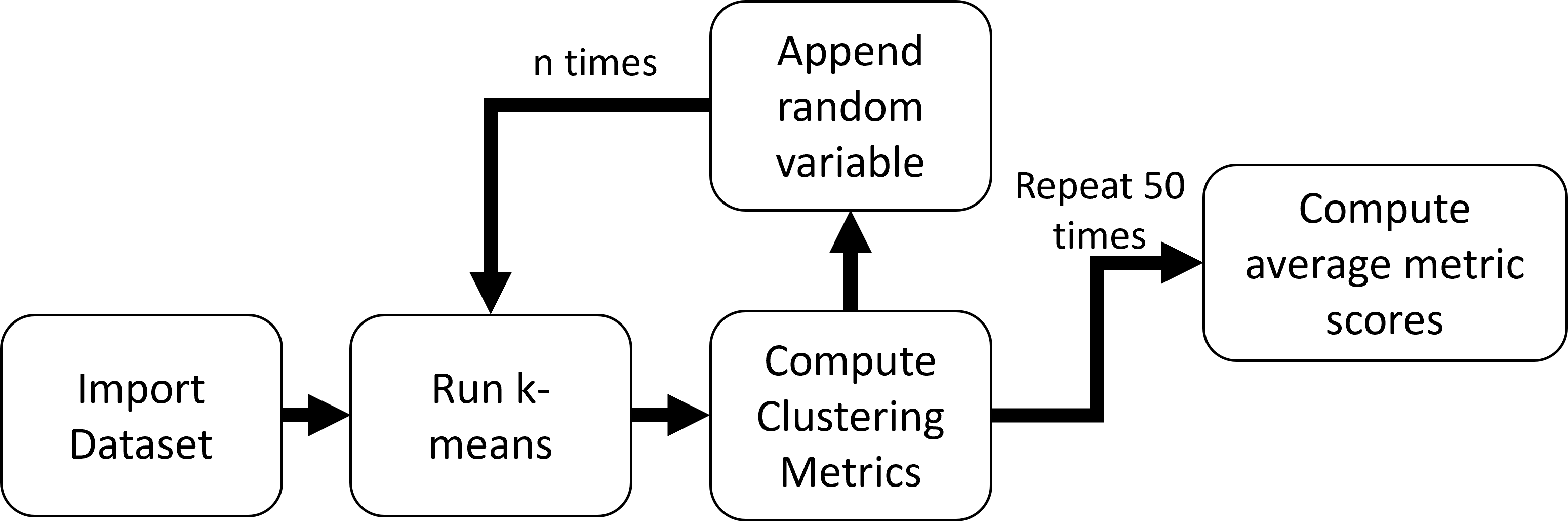}
    \caption{Workflow for the experimentation used in this work. \label{fig:workflow}}
\end{figure}

To assess the impact of random variables on clustering performance we use a set of well defined clusters with associated ground truth labels outlined in Section~\ref{sec:data} as our baseline. For each dataset we iteratively append one random variable to each instance in the dataset, increasing its dimensionality by one each time. Within the iteration cycle, we perform k-means clustering on the data and evaluate the results with the metrics outlined in Section~\ref{sec:metrics}. The experimental workflow is summarised in Fig.~\ref{fig:workflow}. We then obtain the distribution of each metrics as a function of additional random variables. We represent the additional variables as a ratio of random variables to `real' features in the data for comparison across different dataset dimensions. This will highlight the dependence of clustering performance on the proportion of random variables in the data.

Since we have the ground truth we know how many clusters there are in the datasets. Moreover, as the clusters are well defined and separated, we can compare clustering performance to this baseline, and therefore attribute any difference directly to the inclusion of the random variables. As we are iteratively increasing the dimensionality by adding random variables, and have several baseline datasets of increasing dimensionality  of `real' features (all of which well defined clusters), we are also able to examine general properties such as proportion of random numbers. 

We generate random numbers using the mean ($\mu$) and standard deviation ($\sigma$) of the features in the original data. We compare the effect of random numbers generated from a Gaussian distribution, $R_G$
\[
  R_G \sim \mathcal{N}(\mu_r,\,\sigma_r^{2}),
\] 
where 
\[ 
\mu_r = \sign(\mu+\sigma)\eta~,
\]
\[
\sigma_r = \sigma(1\sign\eta)~.
\]
Here $\eta$ is a random number $\left[0,1\right)$ and $\sign$ represents the sign ($\pm$) determined by an additional random number, $\mathcal{R} \in \left[0,1\right)$,
\[
    \sign = 
\begin{cases}
    + , & \text{if } \mathcal{R}\geq 0.5\\
    - , & \text{otherwise}
\end{cases}~.
\]
These are generated for each random value added hence, for the Gaussian distributed values, each random value added has a different mean ($\mu_r$) and standard deviation ($\sigma_r$). We also consider random variables generated from a uniform distribution for identifying any impact due to noise distribution. For the uniform distributed random variable, $R_U$, we sample random numbers for the range $\left[ -\left(\mu + 2\sigma \right), +\left(\mu + 2\sigma \right)  \right]$

Finally, as scaling has been shown to affect clustering results,
~\cite{leonard_kaufman_peter_j_rousseeuw_introduction_1990} we also examine these effects in our experiments. We compare unscaled data, generated with the distributions outlined above, with popular scaling methods. Specifically we consider {\emph Centered} data, where each variable has the mean subtracted yielding a mean of zero in the scaled data, and {\emph Standardized Centered} data, where the variables are centered and scaling to have unit variance. Through the rest of the article we refer to these scaling as `Centered' and `Standardized' respectively.

We perform k-means with $k$ specified from the ground truth labels in the data, 16 in our case. As our choice of implementation is the popular \emph{k-means++}, \cite{scikit-learn}. We repeat each clustering experiments $50$ times for each appended random variable to find an average clustering metrics and provide confidence bounds for these values. We have used Euclidean distance as our measure in all clustering experiments. 

The proportion of added random variables is reported as a ratio of random variable to meaningful features in the baseline dataset. For instance, a ratio of 0:1 represents the baseline datasets in all cases, i.e. no added random variables, whereas a ratio of 2:1 means there are twice as many random variables in the data as there are informative features. In the case of a 2:1 ratio, Dim-$32$ has 64 random variables and 32 informative features, whereas Dim-$128$ has 256 random variables and 128 informative features. This allows comparisons to be made across the varying dimensionality datasets and generalisation of the results.

\section{Results}

\begin{figure*}[h]
    \centering
    \includegraphics[width=0.95\textwidth]{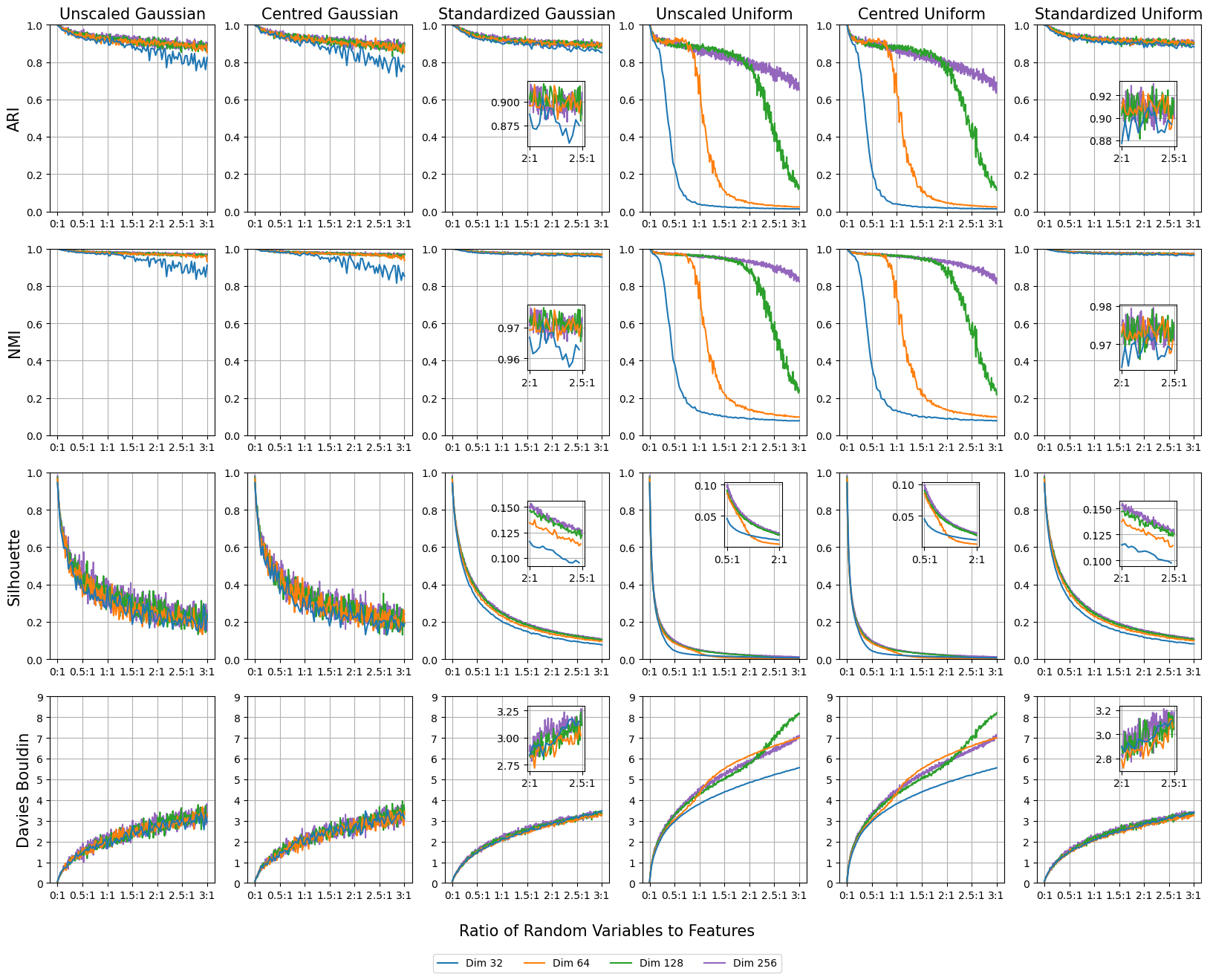}
    \caption{A comparison of clustering performance with different random number generations and data scaling methods. Columns depict different scaling and random number generation methods, and rows illustrate clustering performance metrics averaged over 50 independent runs. The ratio of random variables to informative features ranges from 0:1 (baseline model) to 3:1 where 75\% of the input data is randomly generated and therefore does not correlated to the cluster label. Note that higher Davis-Bouldin scores indicate worse clustering performance unlike the other metrics. \label{fig:metrics}}
\end{figure*}

\begin{figure*}[t]
    \centering\includegraphics[width=0.95\textwidth]{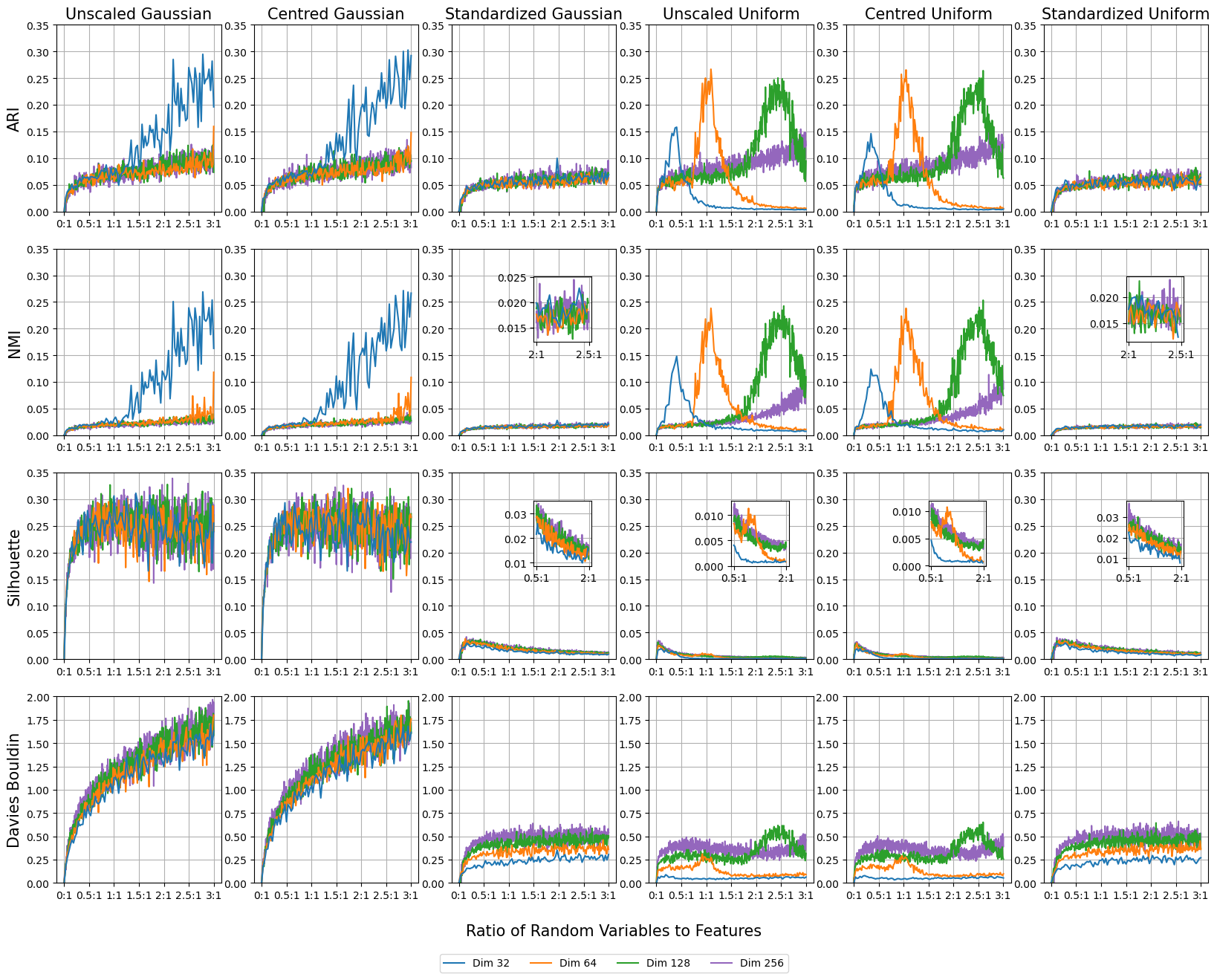}
    \caption{Standard deviation ($\sigma$) value of clustering metrics score for 50 independent repeats. As in  Fig.~\ref{fig:metrics} we plot these values as a function of the proportion of random variables to features. Rows and columns are as in  Fig.~\ref{fig:metrics}. \label{fig:std}}
\end{figure*}

Figure~\ref{fig:metrics} illustrates the results of this work and summarising the dependence of appended random values on clustering performance metric under different scaling methods and when using different distributions to sample the random numbers. For clarity we also look at the standard deviation of these curves in Fig.~\ref{fig:std} which follows the same structure as Fig.~\ref{fig:metrics}.

Overall, we observe the same behaviour in centered and unscaled data for all metrics and both random number distributions. Each of the four dataset demonstrate the same dependence when unscaled or centered, indicating that dimensionality does not influence this. We also see comparable values and dependencies in the standard deviation plots in Fig.~\ref{fig:std} for unscaled and centered data for both random number distributions. 
It is worth reiterating that the appended random variables generated from the Gaussian distribution each have a different mean and variance. As such the unscaled and centered data are different but appear to have the same dependency on appended random variables. Therefore, there is no observable benefit in centering the data when investigating random or uninformative features in the data for unsupervised tasks. For all metrics standardizing the data removes any discrepancy between random variables generated from a Gaussian or Uniform distribution. Moreover, standardizing the data reduces the standard deviation in performance scores in repeated runs for all configurations, with the exception of the Silhouette Coefficient for Uniform random variables which are comparable. 

ARI and NMI behave qualitatively the same across all configurations in Fig.~\ref{fig:metrics}. When considering Gaussian random variables, scaling has little or no effect, with standardized data reducing the larger gradient of Dim-32 compared to the higher dimensional datasets making the dependence on appended random variables comparable across all dimensions. The larger gradient of Dim-32 is accompanied by increasing variation in ARI and NMI with increasing proportions of random variables, see Fig.~\ref{fig:std}. Both metrics are insensitive to large proportions of random variables, indicating high quality clustering performance even when 75 \% of the data is random noise. 
\\~\\
When considering uniform distributed random variables, scaled and centred data exhibit greater sensitivity to added random variables when assessed by ARI and NMI (noted by the steeper initial gradient in the curves in Fig.~\ref{fig:metrics}). However, there is a `tipping point' where the scores rapidly decrease to near zero. The location of this tipping point appears to be dependent on the dimensionality of the baseline data. Higher dimensions exhibit tipping points at higher proportions of random variables to features and the rate of reduction of score appears to be reduced. The start of these tipping points manifests as an abrupt increase in standard deviation (Fig.~\ref{fig:std}). The inflection point in Fig.~\ref{fig:metrics} also corresponds with a maximum in the standard deviation in Fig.~\ref{fig:std}. Standardizing the data removes this dependency completely and resemble the standardized uniform curves closely resemble the standardized Gaussian curves in terms of scores (Fig.~\ref{fig:metrics}) and their variability (Fig.~\ref{fig:std}).
\\~\\
The Silhouette Coefficient and the Davies-Bouldin scores show clear dependence on the proportion of random variables, both showing larger gradients initially before reducing to a lower rate of change as observed in Fig.~\ref{fig:metrics}. The Silhouette Coefficient shows a rapid decrease in score from the baseline before indicating a plateau, this is more noticeable in the Uniform random number data. 
The Davies-Bouldin score has a comparatively lower initial gradient but appears to not to plateau, again more noticeable in the Uniform data. This may be due to the increase in intra-cluster distance with the addition of the random variables in line with the curse of dimensionality. 

Specifically for the Gaussian random variables, standardizing data for both Silhouette Coefficient and Davis-Bouldin metrics reduces the variability seen in the curves in Fig.~\ref{fig:metrics}, see Fig.~\ref{fig:std}. Similarly to the scores, the standard deviation plateaus for the Silhouette Coefficient and continues to increase for Davis-Bouldin. When using standardized data, the Silhouette Coefficient yield scores at the lower end of the range seen in the unscaled and centered data, indicating lower quality clustering with increasing number of random variables added. Whereas, for the Davis-Bouldin metric, standardized data appear to have comparable values to the unscaled and centered data. For both metrics, standardized data drastically reduces the variability between runs (Fig.~\ref{fig:std}), and in the case of the Davis-Bouldin the standard deviation plateaus. 

For the Uniform random variables, both Silhouette Coefficient and Davis-Bouldin show rapid degradation of score with increasing proportions of random variables. This dependence is stronger, and exhibit much lower variation, than in the Gaussian case. For unscaled and centered data, both metrics also show some subtle structure in the curves that is dependent on the dimensionality of the baseline data. This results in clustering performance being worse for lower dimensional data at certain proportions of random variables. For example 1.5:1 Dim-64 has a higher Davis-Bouldin score than Dim-128 and Dim-256, whereas at 2.5:1 the Dim-128 curve has overtaken Dim-64 and Dim-256. At 3:1 it appears that Dim-256 is higher than Dim-64 and may exhibit the same pattern, though this is outside of the range of our analysis. This dependence on dimensionality manifests as peaks in the standard deviation plots in Fig.~\ref{fig:std}, albeit much smaller peaks than seen in ARI and NMI. 

The Silhouette Coefficient shows similar patterns but these are obfuscated by the dynamic range of these curves and are visible in the insets in Fig.~\ref{fig:metrics} and Fig.~\ref{fig:std}. Standardizing the data removes this structure for both metrics and leads to curves resembling the standardize Gaussian data.

\section{Conclusion}

These results indicate that the Silhouette Coefficient and the Davies-Bouldin score are the most sensitive to irrelevant features in all cases. The Silhouette Coefficient exhibits rapid decrease in value in response to comparatively low levels of added irrelevant features indicating it is the most sensitive from a {\em perfect} feature set baseline. The Davies-Bouldin score also exhibits a rapid increase when irrelevant features are added, and its trend implies that it will not plateau. Both the Silhouette Coefficient and Davies-Bouldin score provide useful measurements of cluster quality that are sensitive to the addition of irrelevant features. As these metrics do not require ground truth labels, they are well suited as objective functions to optimise in feature selection for unsupervised tasks with unknown amounts of irrelevant features or when a {\em perfect} feature baseline is not available. 

Conversely, ARI and NMI show a resilience to irrelevant features. For Uniform random numbers, this resilience is up to a critical point, that appears to be dependent on the dimensionality of the data. No critical points were observed with Gaussian random numbers and ARI and NMI maintained very high scores even at high proportions of irrelevant features relative to informative features in the data. This indicates that these metrics may not be useful for evaluating the clustering of noisy data, particularly if the noise is Gaussian distributed.

Finally, we also observe that Standardized data reduces the variability of the clustering results between runs and also provides comparable results between Gaussian and Uniformly distributed random variables. It also removes the appearance of tipping points in the Uniform random numbers that is dependent on the dimensionality of the baseline data. 


\bibliographystyle{unsrt}
\bibliography{ClusterSensitivitylib}


\section*{Contribution of authors}
Spencer Thomas designed the study and experiments. Miles McCrory and Spencer Thomas implemented and conducted all experiments. All authors interpreted the results and prepared the manuscript. The authors would like to thank Sam Bilson and Peter M Harris (NPL) for useful feedback on the manuscript. 

\section*{Sources of funding}
This work was funded by the Department for Science, Innovation and Technology through the National Measurement System.  




\end{document}